\documentclass[conference]{IEEEtran}
\IEEEoverridecommandlockouts
\usepackage{cite}
\usepackage{amsmath,amssymb,amsfonts}
\usepackage{algorithmic}
\usepackage{graphicx}
\usepackage{textcomp}
\usepackage{xcolor}
\usepackage{url}
\usepackage{booktabs}
\usepackage{multirow}
\usepackage{makecell}
\usepackage{arydshln}
\usepackage{float}
\usepackage{comment}
\usepackage{tabularx}

\def\BibTeX{{\rm B\kern-.05em{\sc i\kern-.025em b}\kern-.08em
    T\kern-.1667em\lower.7ex\hbox{E}\kern-.125emX}}

\begin{document}

\title{Enhancing Action Recognition by Leveraging the Hierarchical Structure of Actions and Textual Context}

\author{
Manuel Benavent-Lledo$^{a,*}$, David Mulero-Perez$^{a}$, David Ortiz-Perez$^{a}$,\\ Jose Garcia-Rodriguez$^{a}$ and Antonis Argyros$^{b}$\\
$^{a}$Department of Computer Technology, University of Alicante, Alicante, Spain
\\$^{b}$Institute of Computer Science, FORTH, Heraklion, Crete, Greece\\
$^*$Corresponding author: {\tt mbenavent@dtic.ua.es} \\}

\maketitle

\begin{abstract}
We propose a novel approach to improve action recognition \emph{by exploiting the hierarchical organization of actions and by incorporating contextualized textual information, including location and previous actions, to reflect the action's temporal context}. To achieve this, we introduce a transformer architecture tailored for action recognition that employs both visual and textual features. Visual features are obtained from RGB and optical flow data, while text embeddings represent contextual information. Furthermore, we define a joint loss function to simultaneously train the model for both coarse- and fine-grained action recognition, effectively exploiting the hierarchical nature of actions. To demonstrate the effectiveness of our method, we extend the Toyota Smarthome Untrimmed (TSU) dataset by incorporating action hierarchies, resulting in the \emph{Hierarchical TSU dataset}, a hierarchical dataset designed for monitoring activities of the elderly in home environments. An ablation study assesses the performance impact of different strategies for integrating contextual and hierarchical data. Experimental results demonstrate that the proposed method consistently outperforms SOTA methods on the Hierarchical TSU dataset, Assembly101 and IkeaASM, achieving over a 17\% improvement in top-1 accuracy.

\end{abstract}


\section{Introduction}\label{sec:intro}
Recognizing actions in video sequences has become a fundamental task in a wide range of real-world applications, particularly in assistive technologies that enhance safety, efficiency, and quality of life. These technologies are increasingly integral to diverse domains, including video surveillance~\cite{oad-surveillance}, autonomous driving~\cite{Kim_2019_CVPR,ramanishka2018CVPR}, and human-robot interaction~\cite{oad-robot}. Specifically, video monitoring systems, such as smarthomes and robots, stand to benefit immensely from these advances. These systems have the potential to significantly improve the quality of life for older adults in domestic environments and enhance worker safety in industrial settings, thereby driving innovation in both healthcare and industrial safety.

In some of these cases, the effectiveness of the provided solutions depends on their ability to operate online and in real-time. In other scenarios, however, offline analysis of behavior may be sufficient. In such cases, considering the broader context provided by the entire video can improve model performance. For example, routine analysis can be used to detect behavioral changes in elderly individuals~\cite{TALAVERA2020107330,MENCHON2023101846}, or to identify errors in industrial assembly tasks~\cite{Flaborea_2024_CVPR,Sener_2022_CVPR,Schoonbeek_2024_WACV}.

In the offline setting, action recognition and action detection emerge as the primary tasks in analyzing trimmed and untrimmed videos, respectively. Action recognition involves categorizing videos into predefined classes~\cite{wang2021actionclip,wu2023revisiting}, whereas action detection extends this by identifying the temporal boundaries of each action within the video~\cite{dai2022ms,dai2021pdan}. While the former pursues a straightforward goal, the latter benefits from the sequential nature of actions to enrich the temporal context, thereby reinforcing the robustness of the employed model and solution.

A key aspect of both tasks is the granularity of the action categories, which can significantly affect the performance and applicability of the model. The definition of the different action categories is typically left to the discretion of the dataset authors, resulting in varying levels of abstraction. For example, the task of \emph{make coffee} may be decomposed into several steps involving different objects: \emph{open coffee maker, place filter}, etc. While the coarse granularity is sufficient in some cases, certain applications may require additional refinement to achieve specific goals. For example, robotics applications often require more detailed annotations, including per-joint instructions.

This work presents an action recognition architecture that exploits both the hierarchical structure and the sequential execution of actions. To this end, we adopt a transformer encoder built on the basis of state-of-the-art methods~\cite{Wang_2021_ICCV,ulhaq2022vision,Chen_2022_WACV} that use this architecture to model temporal dependencies. By exploiting the self-attention mechanism inherent in transformer architectures, this model exceeds previous methods such as 3D CNNs~\cite{8099985,sgs2021,10.1007/978-3-030-68238-5_48} and RNNs~\cite{LI201841,he2021db,An_2023_ICCV}, which often face challenges in capturing long-range temporal dependencies.

\sloppy
The Transformer architecture~\cite{vaswani2017attention}, originally designed for natural language processing, has undergone numerous improvements in recent years~\cite{brown2020language,touvron2023llama,devlin2018bert}. These improvements have enabled the models to handle temporal dependencies more effectively and to capture contextual information within textual data. We explore the benefits of these models to incorporate contextual data for action recognition in two different ways. First, by evaluating the effect of rephrasing to enrich textual descriptions of past actions along with camera location information which is achieved by leveraging instructional models such as GPT-3.5~\cite{brown2020language} and Llama~3~\cite{dubey2024llama3herdmodels}. Second,  by encoding textual data into feature vectors using transformer text encoders~\cite{devlin2018bert,sanh2019distilbert}. By fusing contextual information in the form of text embeddings with visual features prior to classification, we improve the overall representation of the ongoing action, enabling more accurate action recognition.

Our model is trained with the dual objective of both coarse-grained and fine-grained action classification. The rationale behind this approach is rooted in the idea that learning both coarse-grained and fine-grained actions together facilitates the recognition of both. Previous studies~\cite{suris2021hyperfuture,10.1007/978-3-031-21062-4_39} have explored action hierarchies under the premise that hyperbolic geometry naturally encodes hierarchical structures. Other works~\cite{10222914, zhao2023antgpt} exploit the goal of the video, \emph{i.e.} a coarse-grained action, to improve action anticipation performance. In our method, we propose a joint loss function to allow the model to better discriminate fine-grained actions. 

For experimentation, we extend the existing Toyota Smarthome Untrimmed (TSU) dataset~\cite{Dai_2022_PAMI}, incorporating action hierarchies into Activities of Daily Living (ADL). The resulting dataset, the Hierarchical TSU dataset, is, to the best of our knowledge, the first hierarchical dataset for ADL. Additionally, we evaluate the proposed method on the Assembly101~\cite{Sener_2022_CVPR} and IkeaASM~\cite{Ben-Shabat_2021_WACV} action recognition benchmarks to provide further insights on the benefits of incorporating contextual information and action hierarchies.

Our implementation and the extended annotations for the Hierarchical TSU dataset are publicly available on Github\footnote{\url{https://github.com/3dperceptionlab/HierarchicalActionRecognition}}.

In summary, the contributions of this paper are the following:
\begin{itemize}
\item We present the Hierarchical TSU dataset, an extension of the original TSU dataset~\cite{Dai_2022_PAMI}, that incorporates coarse-grained annotations and contextual information. Additionally, we compare various strategies for generating textual descriptions of contextual data by leveraging previous actions and camera location.
\item We introduce a novel vision-language transformer architecture for action recognition that leverages hierarchical and contextual information, leading to more accurate action classification. Our approach outperforms state-of-the-art action recognition models in Assembly101 and IkeaASM benchmarks and serves as a strong baseline for the Hierarchical TSU dataset.
\item We conduct extensive experiments to demonstrate the effectiveness of our approach. In addition to outperforming existing methods, we present detailed ablation studies that analyze the contributions of each component, as well as the impact of hierarchical and contextual information, including the role of large language models and the influence of the number of past actions, and of location data.
\end{itemize}

The remainder of the paper is structured as follows.
Section~\ref{sec:rel-work} summarizes the latest related work. In Section~\ref{sec:method} we describe in detail the proposed method. The experiments and results, including details on the Hierarchical TSU dataset, are reported in Section~\ref{sec:exp}, followed by the ablation study in Section~\ref{sec:ablation}. Finally, the main conclusions of this work are discussed in Section~\ref{sec:conc}.
\section{Related Work}\label{sec:rel-work}
We present an overview of pertinent literature concerning action recognition, focusing on the integration of language models and hierarchical structures of actions as tools to enhance action recognition performance.

\subsection{Action Recognition}
Action recognition involves the classification of the action class performed within a video clip. Over time, various methodologies have been devised, with 2D CNN methods relying on single-frame human-object interaction being deemed the least effective due to their lack of temporal context~\cite{9892910}. Temporal modeling stands as a crucial aspect of action recognition. While recurrent neural networks and 3D CNN based methods held sway for several years~\cite{8099985,sgs2021,10.1007/978-3-030-68238-5_48,LI201841,he2021db,An_2023_ICCV,Feichtenhofer_2019_ICCV}, the advent of video transformers~\cite{Wang_2021_ICCV,ulhaq2022vision,Chen_2022_WACV} has significantly enhanced temporal modeling capabilities.

Two primary categories of architectures emerge. The first one comprises end-to-end models such as ViVit~\cite{arnab2021vivit} which uses a pure transformer architecture for video classification inspired by the advances in the image domain. Embeddings are extracted as non-overlapping tubelets that span both the spatial and temporal dimensions. In TimeSformer~\cite{gberta_2021_ICML}, a convolution-free approach was presented with a detailed study on self-attention schemes. Results from this study suggest that divided attention for spatial and temporal features leads to the best performance. Video~Swin~\cite{liu2022video} explores the inductive bias of locality in video transformers by adapting the Swin transformer designed for the image domain~\cite{Liu_2021_ICCV}. MViT~\cite{fan2021multiscale} employs multiscale feature hierarchies with a pyramid of feature activations, allowing effective modeling of simple low-level, and complex high-level visual information. MViTv2~\cite{li2021improved} enhances its predecessor with decomposed relative positional embeddings and residual pooling connections. 

The second category comprises temporal modeling transformers that use pre-trained feature extractors from large datasets such as Kinetics-400~\cite{8099985} and ImageNet~\cite{5206848}. OadTR~\cite{Wang_2021_ICCV} focus on temporal modeling using decoded RGB frames and frozen frame-level feature extractors. In addition, optical flow is computed on the RGB data to improve accuracy. Similarly, TIM~\cite{chalk2024tim} uses frozen visual and audio encoders for feature extraction. Features include a timestamp provided by a Time Interval MLP, so that the model can be queried about the events at a given interval in a specific modality. Other approaches incorporate the vision transformer, ViT~\cite{dosovitskiy2020image}, to fine-tune the feature extractor on the corresponding datasets~\cite{wang2021actionclip, wu2023revisiting,wu2023transferring}, yielding better results with a remarkable increase in terms of computing cost.

Feature extractors usually consist of 2D or 3D CNNs such as ResNet~\cite{he2016deep}, InceptionV3~\cite{szegedy2016rethinking} or I3D~\cite{8099985}. On the contrary, MM-Vit~\cite{Chen_2022_WACV} diverges by operating on multimodal features extracted from compressed videos, including I-frames, motion vectors and audio features. Similar strategies are observed in~\cite{ming2023fsconformer,wang2022deformable,MING2024127389}.

\subsection{Language Models for Action Recognition}
The remarkable capabilities of large language models in temporal modeling and feature representation have been thoroughly investigated in recent years~\cite{brown2020language, touvron2023llama, devlin2018bert}. These capabilities have found application in various vision tasks, notably in image and video captioning. These approaches focus on generating textual descriptions of videos~\cite{8627985}. Recenttly, research has explored the use of language models to improve action recognition results. For example, ActionCLIP~\cite{wang2021actionclip} uses a contrastive learning approach inspired by CLIP~\cite{radford2021learning}. Instead of image captions, action classes are converted to prompts, which are then compared to the aggregated representation of a video. Spatial features from frames are extracted from a fine-tuned version of CLIP's visual encoder, based on ViT~\cite{dosovitskiy2020image}. Similarly, VideoCLIP~\cite{xu2021videoclip} extends this learning paradigm to various video understanding tasks. Building on these advances, Text4Vis~\cite{wu2023revisiting, wu2023transferring} proposes to initialize a frozen classifier for action recognition using text embeddings derived from language models. Moreover, BIKE~\cite{bike} introduces a novel framework that facilitates bi-directional cross-model knowledge transfer from vision to language models, with the aim of improving action recognition in videos.

Furthermore, the capabilities of language models in capturing temporal dependencies and contextual understanding have been exploited to model past actions as well~\cite{CAESAR2024100072}. VLMAH~\cite{manousaki2023vlmah} presents a visual-linguistic approach to modeling action history that is particularly useful for instructional videos due to the sequential nature of the actions. Similarly, Furnari and Farinella~\cite{furnari2020rolling} use ``rolling-unrolling'' LSTMs to succinctly summarize past actions. In contrast, AntGPT~\cite{zhao2023antgpt} exploits large language models for in-context learning~\cite{brown2020language} long-term action anticipation, \emph{i.e.} by providing few ground-truth examples. Similarly, Zhang et al.~\cite{10484396} explore the benefits of incorporating object representations for this task. Encoded representations of cropped objects from RGB frames, along with bounding boxes and object labels encoded as text embeddings, are shown to improve the model's performance.

\subsection{Action Hierarchies}
Hierarchical structures have been explored in various ways for action understanding tasks, such as modeling features at multiple levels~\cite{dai2021pdan,9340987,Richard_2017_CVPR,mcaf2025} or leveraging annotations with varying granularity. In this work, we focus on the latter to investigate the potential of structured datasets for improving action understanding.

While current action recognition techniques have made significant progress, they often fall short in segmenting actions into distinct phases, which is required for many real-world applications. To address this gap, the authors of the FineGym dataset~\cite{shao2020finegym} introduced a sports video dataset with a three-level semantic hierarchy. Previous research~\cite{suris2021hyperfuture,10.1007/978-3-031-21062-4_39} has explored this dataset using the premise that hyperbolic geometry inherently encodes hierarchical structures. Using the Poincaré ball, these studies define a distance metric between predictions and observations, where points closer to the center of the ball represent abstract embeddings, while those near the edge denote specific ones. Essentially, edge proximity indicates higher model confidence.

Alternatively, Timeception~\cite{Hussein_2019_CVPR} redefines the notion of activity, restricting it to ``complex actions'' characterized by: (1)~composition - consisting of several simpler actions, (2)~temporal order of these actions, and (3)~extent - recognizing the variability in temporal length between actions. By introducing the Timeception layer, the architecture tolerates both long-range temporal dependencies and variations in the temporal extent.

The previously introduced AntGPT~\cite{zhao2023antgpt} also benefits from hierarchical information by incorporating goal information extracted from past actions using a large language model. A large language model is used to extract these goals in the form of text embeddings, which are then fused with visual observations to enhance long-term action anticipation. To generate hierarchical labels, AntGPT relies on twelve in-context examples, either manually curated or pseudo-labeled using large language models applied to video titles or descriptions. Similarly, in~\cite{10222914}, authors exploit the goal concept to improve action anticipation in industrial scenarios by introducing a consistency loss to ensure alignment between coarse-grained and fine-grained predictions.

The industrial domain has particularly benefited from hierarchical annotations~\cite{Sener_2022_CVPR,Ben-Shabat_2021_WACV}, which may be explicitly provided or derived by defining coarse-grained related goals (e.g., assembling or disassembling a product). Other domains, such as autonomous driving, have also explored hierarchical annotations to capture the progression from high-level driving maneuvers to specific low-level actions~\cite{Martin_2019_ICCV}.

Despite these advances, it is important to note the scarcity of annotated hierarchical data in the ADL domain, which prompts researchers in~\cite{suris2021hyperfuture} to manually extend existing datasets for experimentation. Alternatively, synthetic data generators, such as Virtualhome~\cite{virtualhome}, offer a hierarchical approach to generating video data, providing a valuable resource to address this limitation.
\section{Hierarchical TSU Dataset}\label{sec:dataset}
\begin{figure*}[t]
    \centering
    \includegraphics[width=\textwidth]{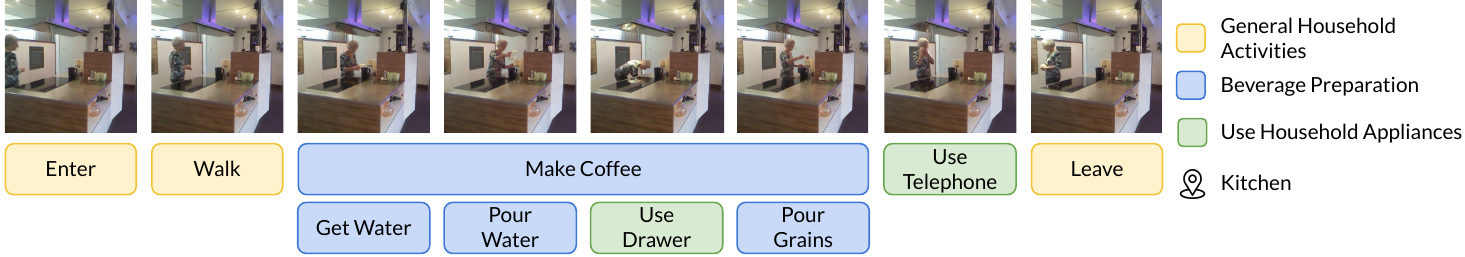}
    \caption{\textbf{Hierarchical annotation example in the TSU dataset.} Video frames are annotated with fine-grained action labels, including composite activities (e.g., \emph{Make coffee}). Each color corresponds to one of the coarse-grained action categories.}
    \label{fig:hierarchies}
\end{figure*}

We introduce the Hierarchical TSU dataset, an extension of the Toyota Smarthome Untrimmed dataset~\cite{Dai_2022_PAMI}, designed for the study of action hierarchies in activities of daily living. This dataset aims to incorporate and evaluate two complementary sources of information for action recognition: (1)~contextual cues, such as prior actions, spatial location, and scene context, and (2)~hierarchical organization of actions, reflecting the natural structure of human behavior.

We adopt the TSU dataset as the foundation for our work due to several key advantages. Unlike other ADL datasets (e.g., Toyota Smarthome~\cite{9008135} or ETRI-Activity~\cite{jang2020etri}), TSU provides temporally ordered annotations of actions, static camera viewpoints with known spatial locations, and a comprehensive taxonomy of fine-grained action labels. Specifically, it includes 51 distinct action classes recorded from 7 fixed cameras in various rooms of a house. The dataset features 536 untrimmed videos, each averaging 21 minutes in duration, and captures daily activities performed by 18 elderly individuals aged 60 to 80. This diversity in subject demographics and household settings makes TSU particularly well-suited for hierarchical and context-aware action recognition. In total, TSU contains over 64,000 annotated fine-grained action segments, which correspond to the trimmed action segments that compose the Hierarchical TSU dataset.

Among the 51 annotated actions are elementary activities such as \emph{walk} and \emph{get water}, as well as composite actions like \emph{make coffee} and \emph{cook}, which often encompass or temporally overlap with elementary actions. Although these composite and elementary actions exhibit clear hierarchical relationships, we preserve the original flat annotation structure of the TSU dataset to maintain consistency. Building upon this, we introduce an additional semantic layer by organizing the 51 action classes into a two-level hierarchy. Specifically, we manually assign each fine-grained action to one of seven coarse-grained categories, based on contextual semantics, location of execution, and visual similarity. The resulting high-level categories are: \emph{beverage preparation}, \emph{general household activities}, \emph{cleaning}, \emph{prepare breakfast}, \emph{use household appliances}, \emph{cook}, and \emph{drink}. Table~\ref{tab:htsu-stat} summarizes the distribution of fine-grained actions within each coarse-grained category, as well as the number of annotated segments per class. While our work focuses on action recognition, we note that the introduced hierarchical annotations are also well-suited for action detection tasks within the TSU setup, as they enable richer temporal modeling and more structured predictions.

\begin{table}[t]
\setlength{\tabcolsep}{3pt}
\caption{Distribution of fine-grained actions per coarse-grained category.}\label{tab:htsu-stat}
\footnotesize
\centering
\begin{tabularx}{\linewidth}{@{} >{\hsize=1.7\hsize}X >{\hsize=0.9\hsize}X >{\hsize=0.6\hsize}X >{\hsize=0.8\hsize}X @{}}
\toprule
\textbf{Coarse-Grained } & \textbf{\footnotesize\# Fine-Grained} & \textbf{Samples} & \textbf{Avg. Dur. (s)} \\\midrule
Beverage Preparation & 11 & 2263 & 6.4 \\
General Household Activities & 14 & 40779 & 7.8 \\
Cleaning & 5 & 3457 & 10.1 \\
Prepare breakfast & 5 & 3787 & 7.83 \\
Use Household Appliances & 7 & 5645 & 13.33 \\
Cook & 5 & 6242 & 7.83 \\
Drink & 4 & 5241 & 2.87 \\\bottomrule         
\end{tabularx}
\end{table}

For contextual action recognition, we extract trimmed segments corresponding to individual fine-grained actions from the untrimmed videos. Because composite activities may span or overlap multiple fine-grained segments, the dataset includes overlapping annotations at both hierarchical levels, as illustrated in Figure~\ref{fig:hierarchies}. To enrich the temporal context, we additionally process preceding actions in the sequence, encoding them as textual input. Finally, the known spatial configuration of the cameras, detailed in~\cite{Dai_2022_PAMI}, is leveraged to associate each action with its location, thereby providing valuable spatial context for downstream models.

\section{Methodology}\label{sec:method}
\begin{figure*}[t]
    \centering
    \includegraphics[width=0.8\textwidth]{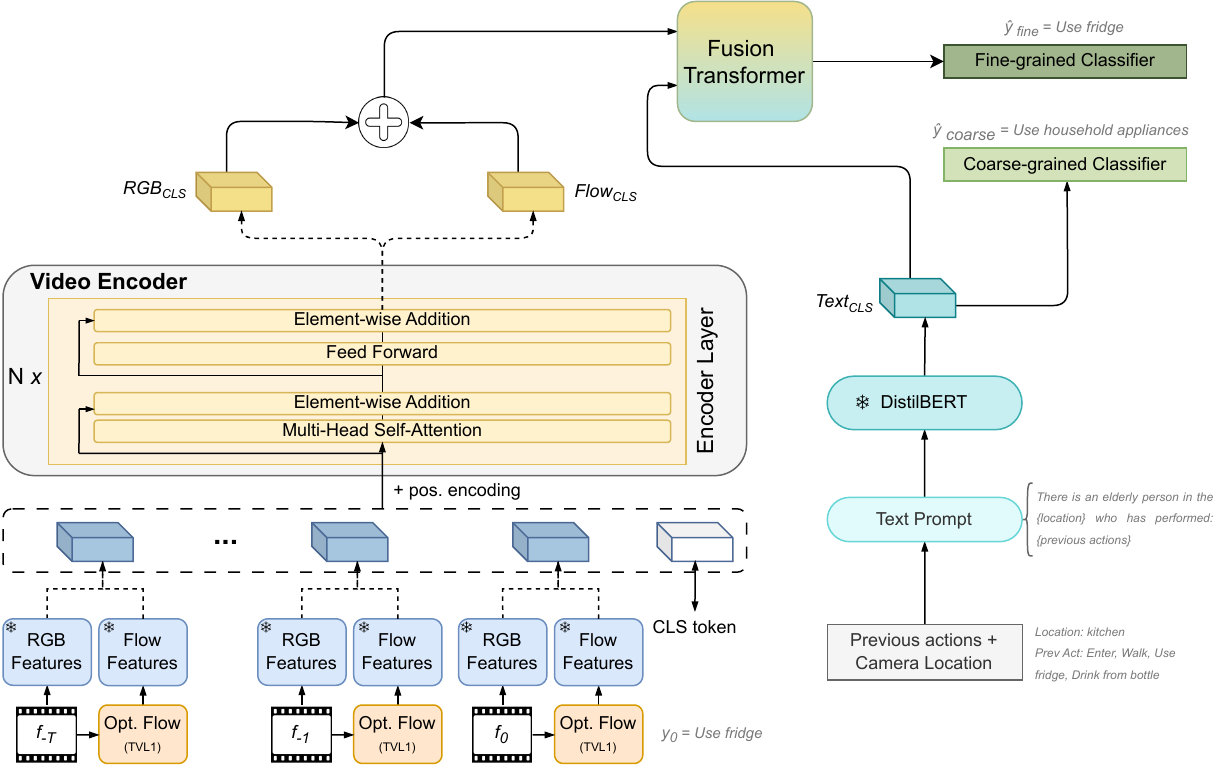}
    \caption{\textbf{Overview of the proposed action recognition architecture.} From a video input, frozen feature extractors obtain spatial features from RGB frames and temporal motion features from optical flow. These features, along with a class token, are processed through individual video transformer encoders for each visual modality, capturing long-range temporal dependencies. Dashed lines indicate the use of either RGB or flow features and their respective embeddings, though only a single video encoder is depicted for simplicity. Additionally, DistilBERT extracts textual features that represent the current location, previous actions and dataset context. Coarse-grained actions are recognized based on contextual information, specifically utilizing the textual embeddings while the fine-grained classifier leverages the fused features from the fusion transformer.}
    \label{fig:arc}
\end{figure*}

We present the proposed method for action recognition using hierarchical action structures and contextual information. Figure~\ref{fig:arc} illustrates an overview of the proposed architecture.

\subsection{Video Encoder}
We adopt a standard Transformer's encoder to model temporal dependencies as in~\cite{Wang_2021_ICCV}. As noted earlier, the self-attention mechanism in transformer architectures has significantly outperformed previous methods based on 3D CNNs and RNNs, especially in the context of temporal modeling for videos. This mechanism allows for more effective capture of long-range dependencies and complex temporal relationships within video sequences.

Given a trimmed video $V = \{f_t\}^{0}_{t=-T}$ associated with a coarse-grained action, and one or more fine-grained actions, the frames $f_t$ are grouped into blocks of size $B$, resulting in $(T+1)/B$ chunks. For each block, we use a frozen feature extractor to obtain spatial features from RGB and optical flow frames, and employ separate video encoders with the same structure for each modality. We use the central frame of the block for RGB features, and the $B$ frames sequence for optical flow features. These features are mapped into a $D$ dimensional feature space, which we can formalize as $F = \{\mathcal{O}_t\}^0_{t=-T} \in \mathbb{R}^{(T+1) \times D}$, where $\mathcal{O}_t$ represents the output token of the transformer encoder at time $t$.

\sloppy
We extend the feature vector with an additional learnable class token with $\mathcal{O}_{CLS} \in \mathbb{R}^D$. This token is used to learn the aggregated representation of the entire input, and is combined with the feature sequence as $\hat{F} = Stack(\{\mathcal{O}_t\}^0_{t=-T}, \mathcal{O}_{CLS}) \in \mathbb{R}^{(T+2)\times D}$. Alternatively, the aggregated representation of the input can be obtained as the mean of the tokens in the input sequence as in~\cite{beyer2022better}.

Due to the lack of frame positional information in the encoder, we additionally embed positional encoding. Experiments comprise fixed sinusoidal inputs and trainable embeddings. We add positional encoding $E_{pos} \in \mathbb{R}^{(T+2)\times D}$ as shown in Equation~\ref{eq:pos-enc}, \emph{i.e.}, an element-wise addition to preserve positional information:
\begin{equation}
    X_0 = \hat{F} + E_{pos}.
    \label{eq:pos-enc}
\end{equation}

The fundamental element of the transformer model is the Multi-head Self-Attention (MSA) mechanism. In essence, self-attention enables each token to engage with others, enhancing its ability to gather valuable semantic insights. This process involves calculating dot products between queries and keys, followed by a softmax function to determine the importance assigned to each value. Formally, 
\begin{align}
    X' = Norm(X_0), \\
    Attention(Q_i,K_i,V_i) = softmax \left ( \frac{Q_iK_i^T}{\sqrt{d_k}} \right ) V_i, \\
    H_i = Attention(Q_i,K_i,V_i),
\end{align}
where $Q_i = X'W_i^q$, $K_i=X'W_i^k$ and $V_i=X'W_i^v$ are linear layers applied to the input sequence, $W_i^q,W_i^k,W_i^v \in \mathbb{R}^{D \times \frac{D}{N_{head}}}$ with $N_{head}$ represent the number of heads, and ${1}/{\sqrt{d_k}}$ is a scaling factor that helps stabilizing the training and speed up convergence. Note that queries, keys and values are all vector representations. The outputs of each head are concatenated and fed into a linear layer as:
\begin{equation}
    \label{eq:head-concat}
    \hat{H} = Stack(H_1,H_2,...,H_{N_{head}}) W_d \in \mathbb{R}^{(T+2) \times D},
\end{equation}
where $W_d$ is a linear projection.

Finally, the output is fed into a two-layer feed-forward network (FFN) with GELU activation. Meanwhile, layer normalization and residual links are also applied. The video encoder formulas can then be summarized as follows:
\begin{align}
    \hat{H} = MSA(Norm(X_0)),\\
    m'_1= \hat{H} + X_0,\\
    m_n = FFN(Norm(m'_{n-1})) + m'_{n-1},\\
    m'_n=MSA(Norm(m_{n-1})) + m_{n-1},
\end{align}
where $n \in N$ is the \textit{n-th} encoder layer with a total of $N$, $m'$ denotes the output of the first element-wise addition in the encoder layer, and $m_N \in \mathbb{R}^{(T+2) \times D}$ is the final feature representation of the last encoder layer. For the remaining of the paper, and for simplicity, we use $m_{CLS} \in \mathbb{R}^D$ to denote the output representation of the class token from an encoder. This notation is extended to $m_{RGB}$ and $m_{Flow}$ to refer to the same token of the respective modality.

\subsection{Contextualized Textual Information}
Large language models have shown exceptional abilities in temporal modeling and context understanding. To leverage these capabilities, we propose an approach that models past information to enrich the contextualization of ongoing actions, thereby improving action recognition performance. Specifically, given the location and $N$ previous actions, we use a prompt (as illustrated in Figure~\ref{fig:arc}) to provide contextual information, including dataset domain (e.g. an elderly person), environment (e.g. camera location) and prior events, thus modeling longer temporal dependencies. From the generated sentences, we extract feature vectors using DistilBERT~\cite{sanh2019distilbert}, a distilled variant of BERT~\cite{devlin2018bert}, which employs deep bidirectional transformer encoders for text comprehension. As for the video encoder, we utilize the class token to represent the entire input sequence, denoted as $m_{Txt}$.

In addition to DistilBERT, we compare its performance against BERT and explore the use of LLMs for rephrasing  to enhance data variability and richness. Section~\ref{sec:ablation} presents results for different setups, including no rephrasing, and rephrasing using GPT-3.5~\cite{brown2020language} and Llama~3~\cite{dubey2024llama3herdmodels}, as well as an analysis of the optimal number of past actions and the influence of location information on contextualization.

\subsubsection{Exploiting Hierarchical Information from Context}
Contextual information derived from past actions and location provides a general understanding of the ongoing action, effectively yielding a coarse-grained representation. Likewise, this coarse-grained action can serve as supplementary supervision, enhancing fine-grained action recognition performance. To leverage this relationship, we introduce a coarse-grained classifier, which, in conjunction with a joint loss function, improves the accuracy of fine-grained action recognition.

\subsubsection{Fusion of Visual and Textual Embeddings}
In addition to their capabilities in temporal aggregation, transformer architectures excel at fusing features from different modalities~\cite{10123038}. To leverage this advantage, we employ a transformer encoder to combine visual and textual embeddings. Specifically, we concatenate RGB and optical flow features, as this approach has shown improved performance compared to feeding them separately into the fusion transformer. The encoder layers are structured similarly to the video encoder, but without the class token or positional encoding. Formally,

\begin{align}
    m_{VIS} = m_{RGB} + m_{Flow},\\
    m_M = FusionTransformer(m_{VIS}, m_{Txt}),
\end{align}
where $m_{VIS} \in \mathbb{R}^{2 \times D}$ represents the combined RGB and optical flow features which are then mapped to the embedding space of the fusion transformer, denoted as $\hat{D}$. The output $m_M \in \mathbb{R}^{2 \times \hat{D}}$ results from the $M$ layer of the fusion transformer. The final representation from the fusion transformer, obtained as the mean of the output tokens, is denoted as $m_{Fus}$.

\subsection{Training}
The model is trained on a dual classification objective. First, a fine-grained classifier uses $m_{Fus}$ as input. Second, a coarse-grained classifier leverages contextual information from $m_{Txt}$. The resulting logits from both classifiers are used for multi-class classification, given the existing overlaps between composite actions, as discussed in the next section. 

The supervision of both fine-grained and coarse-grained actions is performed through a joint loss function, defined as:
\begin{align}
    \mathcal{L} = BCE(\sigma(z_{coarse}), y_{coarse}) + BCE(\sigma(z_{fine}), y_{fine}),
\end{align}
where $z$ and $y$ represent predicted logits and ground truth labels, respectively.  The function BCE denotes the binary cross-entropy loss, and $\sigma$ is the sigmoid activation function applied to the logits. Using this combined formulation ensures numerical stability by applying the sigmoid and binary cross-entropy operations together, rather than separately.

\section{Experiments}\label{sec:exp}
This section presents the experimental setup and results to demonstrate the effectiveness of the proposed method compared to the equivalent visual-only approach, and state-of-the-art methods across different datasets.

\subsection{Experimental Setup}
\noindent\textbf{Evaluation:} We evaluate the results of our experiments on top-$k$ accuracy following previous work on action recognition~\cite{wang2021actionclip,wu2023revisiting,9008135,Sener_2022_CVPR,Ben-Shabat_2021_WACV}. Top-$k$ accuracy measures how often the correct action label appears among the top-$k$ labels predicted by the model. We use the cross-subject evaluation introduced in the TSU dataset~\cite{Dai_2022_PAMI}, which uses 11 subjects for training and the remaining 7 for testing. Unless specified otherwise, in all tables, the best performance is highlighted in bold, and the best performance within each subgroup, if applicable, is underlined.

\vspace*{0.2cm}\noindent\textbf{Implementation details:} After exhaustive ablation experiments to determine the best configuration, we adopt ViT-H~\cite{dosovitskiy2020image} with a patch size of 14 as RGB feature extractor and Inception v3~\cite{szegedy2016rethinking} from the two-stream network TSN~\cite{TSN2016ECCV} for optical flow features. TLV1 algorithm is used to compute optical flow frames. To extract embedding vectors from textual data we use DistilBERT~\cite{sanh2019distilbert}.

The video encoder is composed of 4 attention layers with a single attention head, and an embedding size of 2048. The input vector is supplemented  with learned position encoding and a CLS token. The resulting RGB and optical flow video embeddings are concatenated, and this visual representation is fused with text embeddings using the fusion transformer composed of 2 encoder layers, with 2 attention heads per layer and an embedding size of 768. The best performance is obtained using 32 blocks as input, which corresponds to $6.4$ seconds of video. The prompt for contextual information contains 5 past actions along with location data.

\vspace*{0.2cm}\noindent\textbf{Training details:} The proposed method is implemented using PyTorch, and all experiments are conducted on Nvidia RTX 4090 GPUs. We employ AdamW optimizer with a weight decay of $0.1$ and a base learning rate of $5 \times 10^{-5}$. Models are trained for $100$ epochs with a batch size of $32$, employing an early stopping mechanism with a patience of $20$ epochs. We use gradient clipping to avoid exploding gradients and warm up the learning rate for 5 epochs.  We set the block size to $5$, which implies a downsampling rate of 5 on the 25 fps videos from the TSU dataset.

\subsection{Results}
To determine the best combination of visual input with contextual data that provides the best top-$k$ accuracy, Table~\ref{tab:main-res} presents the results for the best settings obtained after extensive ablation experiments (see Section~\ref{sec:ablation}).

For each experiment, we evaluate using both a fine-grained-only training objective (-), and a dual training objective with the proposed joint loss~($\checkmark$). The results lead to the conclusion that the proposed method outperforms the standard approach relying only on visual inputs. Moreover, for every input combination it is observed that predicting both fine- and coarse-grained actions results in better performance. Note that the observed decrease in top-$5$ accuracy in some cases is primarily due to the model's high confidence in the top classes.

\vspace*{0.2cm}\noindent\textbf{Impact of optical flow on video analytics:} Optical flow is considered an auxiliary modality that complements RGB data, providing a notable performance boost of $1.97\%$ when used in combination with RGB, and even outperforming in the uni-modal approach ($1.26\%$). However, its effectiveness diminishes when combined with contextual information and underperforms compared to the corresponding RGB-based method. When all three modalities (RGB, optical flow, and contextual information) are combined, optical flow contributes only a marginal improvement of $0.15\%$.
  
\vspace*{0.2cm}\noindent\textbf{On the effect of the joint loss function:} Using a joint loss function to predict both fine-grained and coarse-grained actions during training enhances performance across all input modalities. This approach yields up to a $1.37\%$ improvement when fusing RGB and text embeddings. Notably, the top-$k$ performance of the coarse-grained classifier correlates with improvements in fine-grained action recognition.

\vspace*{0.2cm}\noindent\textbf{On the effect of exploiting contextual data:} Contextual data, which includes information about previous actions and location in text format, is dependent on visual observations from past frames. As a result, it is always combined with visual modalities and not used independently. Our results demonstrate that incorporating contextual data significantly enhances performance, yielding a $17.12\%$ improvement compared to the RGB-only fine-grained approach, and outperforming all other fusion combinations.

To further validate the efficacy of the proposed method, we evaluate the best-performing model using contextual data extracted from its own predictions instead of ground truth annotations. Table~\ref{tab:main-res2} presents the top-$k$ accuracy for both fine- and coarse-grained actions. The results demonstrate the robustness of the method, even when relying on inference results. Compared to the previous joint loss results without contextual data, our method achieves a $2.08\%$ improvement with RGB data, a $2.03\%$ improvement with optical flow data, and a $2.92\%$ improvement with the combined RGB and optical flow approach.

\begin{table}[t]
\caption{Comparison of results: input modalities and joint loss on the best settings.}
\label{tab:main-res}
\centering
\begin{tabularx}{\linewidth}{cccccc}
\Xhline{1pt}
\multirow{2}{*}{{\bf Modalities}} &  \multirow{2}{*}{{\bf Joint Loss}} & \multicolumn{2}{c}{{\bf Fine-Grained}} & \multicolumn{2}{c}{{\bf Coarse-Grained}} \\
& & {\bf Top-1} & {\bf Top-5} & {\bf Top-1} & {\bf Top-5} \\\Xhline{1pt}
\multirow{2}{*}{RGB}               & - & 37.83 & 72.69 & -     & -     \\
                               & \checkmark & \underline{38.27} & 73.55 & 75.62 & 99.41 \\\Xhline{0.5pt}
\multirow{2}{*}{Flow}              & - & 39.09 & 72.69 & -     & -     \\
                               & \checkmark & \underline{39.39} & 69.95 & 75.18 & 99.07 \\\hline
\multirow{2}{*}{RGB + Flow}        & - & 39.90 & 72.66 & -     & -     \\
                               & \checkmark & \underline{40.24} & 74.21 & 75.44 & 99.37 \\\hline
\multirow{2}{*}{RGB + Text}        & - & 53.43 & 84.03 & -     & -     \\
                               & \checkmark & \underline{54.80} & 84.37 & \textbf{75.77} & 99.37 \\\hline
\multirow{2}{*}{Flow + Text}       & - & 50.83 & 79.44 & -     & -     \\
                               & \checkmark & \underline{50.91} & 81.07 & 75.29 & 99.56 \\\hline
\multirow{2}{*}{RGB + Flow + Text} & - & 53.98 & 84.36 & -     & -     \\
                               & \checkmark & \textbf{54.95} & 84.11 & 73.77 & 99.15\\\Xhline{1pt}
\end{tabularx}
\end{table}

\begin{table}[t]
\centering
\caption{Action recognition performance when contextual data are obtained from predictions.}
\label{tab:main-res2}

\begin{tabularx}{\linewidth}{
  @{}
  >{\hsize=2\hsize}X   
  *{4}{>{\hsize=0.5\hsize}X} 
  @{}
}
\Xhline{1pt}
\multirow{2}{*}{{\bf Modalities}} & \multicolumn{2}{c}{{\bf Fine Grained}} & \multicolumn{2}{c}{{\bf Coarse Grained}} \\
 & {\bf Top-1} & {\bf Top-5} & {\bf Top-1} & {\bf Top-5} \\\Xhline{1pt}
RGB + Text & 40.35 & 71.92 & 70.88 & 98.04 \\
Flow + Text & 41.42 & 70.99 & 70.43 & 97.85 \\
RGB + Flow + Text & \textbf{43.16} & 71.21 & 66.54 & 97.81 \\\Xhline{1pt}
\end{tabularx}
\end{table}

\subsection{Comparison with State-of-the-art Methods}
To evaluate the reliability and robustness of our approach, we compare our method with existing state-of-the-art architectures for action recognition. Given that this work introduces the Hierarchical TSU dataset, the results serve as a baseline to evaluate where our method stands relative to existing approaches. To this end, we train state-of-the-art methods for action recognition initialized with pre-trained Kinetics 400~\cite{8099985} weights for $100$ epochs, matching the training duration of our proposed method. 3D CNN models, including VideoResNet~\cite{Tran_2018_CVPR}, SlowFast~\cite{Feichtenhofer_2019_ICCV}, and X3D~\cite{Feichtenhofer_2020_CVPR}, are trained using the Adam optimizer with a learning rate of $10^{-4}$, as this configuration showed improved performance for these architectures. Transformer-based models, consisting of Video Swin~\cite{liu2022video}, MViT~\cite{fan2021multiscale,li2021improved}, TimeSformer~\cite{gberta_2021_ICML} and ViVit~\cite{arnab2021vivit}, are trained using the same hyperparameters as our method for consistency. To provide a fair comparison across architectures, we evaluate performance using input sizes that match those required by the respective state-of-the-art models.

Table~\ref{tab:sota} summarizes the findings, demonstrating that our model outperforms both 3D CNN and transformer-based architectures on the Hierarchical TSU dataset when trained under identical input size and hyperparameter settings. The best performance, achieved using ground truth contextual data, is highlighted in bold, while the second-best results, derived from inference contextual data, are underlined for each number of input blocks.

\begin{table}[t]
\centering
\caption{Comparison to state-of-the-art methods trained on the Hierarchical TSU dataset.}
\label{tab:sota}
\begin{tabularx}{\linewidth}{
  @{}
  >{\hsize=1.6\hsize}X
  >{\hsize=\hsize}c   
  *{2}{>{\hsize=0.7\hsize}X}  
  @{}
}
\Xhline{1pt}
{\bf Model} & {\bf \# Input Blocks}  & {\bf Top-1} & {\bf Top-5} \\\Xhline{1pt}
X3D \cite{Feichtenhofer_2020_CVPR} & \multirow{6}{*}{16} & 30.22  & 67.05 \\
MViT (Base) \cite{fan2021multiscale} & & 33.01 & 70.28 \\
MViTv2 (Small) \cite{li2021improved} & & 32.47 & 70.80 \\
TimeSformer \cite{gberta_2021_ICML} & & 30.87 & 66.53 \\
\underline{Ours} (Predictions) & & \underline{37.98} & 67.55 \\
\textbf{Ours} (Ground Truth) & & \textbf{51.14} & 81.02 \\\hdashline
VideoResNet \cite{Tran_2018_CVPR} & \multirow{9}{*}{32} & 36.72 & 75.70 \\
SlowFast \cite{Feichtenhofer_2019_ICCV}  &  & 36.53 & 68.77 \\
Video Swin (Base)  \cite{liu2022video} &  & 36.42 & 74.18 \\
Video Swin (Small) \cite{liu2022video} &  & 37.05 & 74.32 \\
Video Swin (Tiny) \cite{liu2022video} &  & 36.75 & 70.32 \\
TimeSformer \cite{gberta_2021_ICML} & & 37.31 & 70.32 \\
ViVit \cite{arnab2021vivit} & & 37.76 & 69.62 \\
\underline{Ours} (Predictions) & & \underline{43.16} & 71.21 \\
\textbf{Ours} (Ground Truth) &  & \textbf{54.95} & 84.11 \\
\Xhline{1pt}
\end{tabularx}
\end{table}

\subsection{Evaluation across Datasets}
We further evaluate the proposed method on two additional action recognition benchmarks, IkeaASM and Assembly101. These industrial-like datasets are chosen for their inclusion of key elements relevant to our approach: untrimmed videos that support contextual modeling of prior actions, and hierarchical action structures.

\textbf{IkeaASM}~\cite{Ben-Shabat_2021_WACV}: This dataset focuses on furniture assembly tasks in different settings comprising 33 atomic (fine-grained) actions under 4 high-level (coarse-grained) categories, each consisting of a different furniture type. Unlike the Hierarchical TSU, a single fine-grained class is assigned to multiple coarse-grained classes. We follow the default train/test data split using the \emph{top} camera view to ensure comparability with prior work. The model architecture and feature extractors match the best settings on the Hierarchical TSU. Given the absence of known locations, we exclude location information from the prompt, and adapt the domain information to \emph{``a person building furniture''}. Results are presented in Table~\ref{tab:res-ikea}, following \cite{Ben-Shabat_2021_WACV} we report Top-1/3 accuracy.

\textbf{Assembly101}~\cite{Sener_2022_CVPR}: This large-scale dataset captures the procedural assembly and disassembly of toy vehicles. It includes 1380 fine-grained and 199 coarse-grained action classes. Similar to IkeaASM, the dataset allows a fine-grained action to be associated with multiple coarse-grained categories, as fine- and coarse-grained recognition tasks are treated independently. To establish fine-to-coarse relationships in this dataset, we matched corresponding temporal segments.

We use the default train/test split and employ the provided DINOv2~\cite{oquab2023dinov2} RGB features from the v4 camera for comparison, which yields the best performance in the third-person view. Because the features are pre-extracted, optical flow is not used, otherwise, the model setup remains the same. As with IkeaASM, prompts omit location information, and the domain context is described as “a person assembling a toy.” The results are reported in Table~\ref{tab:res-assembly}.

\begin{table}[t]
\centering
\caption{Ablation and comparison to state-of-the-art methods on IkeaASM dataset. SOTA extracted from \cite{Ben-Shabat_2021_WACV}. Visual: RGB + Flow. Contextual: RGB + Flow + Text.}

\label{tab:res-ikea}
\begin{tabularx}{\linewidth}{
  @{}
  >{\hsize=2.5\hsize}X
  >{\hsize=0.5\hsize}X   
  *{2}{>{\hsize=0.5\hsize}X}  
  @{}
}
\Xhline{1pt}
\textbf{Model} & \textbf{Hier?} & \textbf{Top-1} & \textbf{Top-3} \\\Xhline{1pt}
\multirow{2}{*}{Ours (Visual)} & - & \underline{75.77} & 83.68 \\
 & \checkmark & 75.43 & 84.54 \\\hdashline
\multirow{2}{*}{Ours (Contextual)} & - & 76.98 & 88.49 \\
 & \checkmark & \underline{79.04} & 76.98 \\\midrule
C3D~\cite{tran2015learning} & - & 45.73 & 69.56 \\
P3D~\cite{qiu2017learning} & - & 60.40 & 81.07 \\
I3D~\cite{8099985} & - & 57.58 & 76.72 \\
Multimodal (RGB, Pose, Depth) \cite{Ben-Shabat_2021_WACV}  & - & 64.02 & 81.45 \\
Ours (Contextual from predictions) & \checkmark & \textbf{65.29} & 81.96 \\\Xhline{1pt}
\end{tabularx}
\end{table}

\begin{table}[t]
\centering
\caption{Ablation and comparison to state-of-the-art methods on Assembly101 dataset. SOTA extracted from \cite{Sener_2022_CVPR}. Visual: RGB. Contextual: RGB + Text.}
\label{tab:res-assembly}

\begin{tabularx}{\linewidth}{
  @{}
  >{\hsize=2.5\hsize}X
  >{\hsize=0.5\hsize}X   
  *{2}{>{\hsize=0.5\hsize}X}  
  @{}
}
\Xhline{1pt}
\textbf{Model} & \textbf{Hier?} & \textbf{Top-1} & \textbf{Top-5} \\\Xhline{1pt}
\multirow{2}{*}{Ours (Visual)} & - & 72.98 & 84.25 \\
 & \checkmark & \underline{73.06} & 83.96 \\\hdashline
\multirow{2}{*}{Ours (Contextual)} & - & 81.44 & 89.80 \\
 & \checkmark & \underline{81.89} & 89.41 \\ \hline
TSM \cite{lin2019tsm} & - & 38.3 & - \\
TempAgg \cite{sener2020temporal} & - & 40.5 & - \\
Ours (Contextual from predictions) & \checkmark & \textbf{43.47} & 69.48\\\Xhline{1pt}
\end{tabularx}
\end{table}

Results on both benchmarks demonstrate the effectiveness of the proposed method over prior work (see the second half of Table~\ref{tab:res-ikea} and Table~\ref{tab:res-assembly}). On IkeaASM, our method using contextual information from predictions achieves an improvement of nearly 5\% in Top-1 accuracy over the previous state-of-the-art RGB-based method, and over 1\% compared to a multimodal approach that combines RGB, depth and 3D pose data. On Assembly101, it yields almost a 3\% gain in Top-1 accuracy compared to TempAgg~\cite{sener2020temporal}. The ablation studies in the first half of the same tables further highlight the impact of contextual modeling, showcasing the benefits of leveraging longer temporal cues and introducing domain-specific textual information.

In contrast, incorporating hierarchical structures results in only marginal improvement on Assembly101 and even a slight decrease in performance on IkeaASM. We attribute this drop to label ambiguity. In IkeaASM, fine-grained actions are commonly assigned to multiple, or even all, coarse-grained categories, making the hierarchy noisy and less informative. As previously discussed, the hierarchical structure in IkeaASM was not designed explicitly for hierarchical action recognition but rather to organize the data by furniture type. Similarly, in Assembly101, hierarchical labels are treated as a separate task rather than an intrinsic semantic structure. Consequently, instead of guiding the model, these hierarchies can introduce noise and hinder fine-grained recognition. This effect is more pronounced in IkeaASM and only mildly beneficial in Assembly101, where over 50\% of coarse-grained classes share at least one fine-grained action, and only 25\% of fine-grained actions are uniquely assigned. These results underscore the importance of a well-structured and semantically meaningful label hierarchy in action understanding datasets, as unclear or overlapping hierarchies can negatively impact performance.

\subsection{Computational Analysis}
We conduct a comparative analysis of the computational characteristics with respect to the previously introduced baseline methods in Table~\ref{tab:comp-analysis}. Compared to prior works, our model shows a larger learning capacity, reflected in a higher number of trainable parameters, while maintaining a lower inference cost. Methods based on CNN architectures have lighter computational requirements, but they generally under-perform compared to transformer-based approaches.
In contrast, our method achieves substantially better accuracy  than other transformer-based models while requiring less computations during inference. Specifically, our model comprises 143.06 million trainable parameters, augmented by an additional 86.39 million parameters from RGB feature extraction, 10.3 million from optical flow feature extraction and 66.36 million from the transformer textual encoder. These parameters were obtained using the feature extractors that performed best.

In terms of inference complexity, our approach requires 13.78 GFLOPs for the multimodal fusion and classification. The visual feature extraction cost is calculated as $(33.74+2.85) \times 32$ GFLOPs, accounting for both the RGB and optical flow pathways across 32 frames.  Additionally, the textual encoding incurs a cost of 11.18 GFLOPs. Overall, this leads to a substantially lower computational cost compared to transformer-based methods like TimeSformer and ViVit, or CNN-based approaches such as SlowFast, which process a larger number of frames or use additional frames to capture richer temporal cues despite similar input window sizes.

\begin{table}[t]
\caption{Computational requiremments of state-of-the-art methods  (\textit{S: Small, B: Base}).}
\centering
\label{tab:comp-analysis}

\begin{tabularx}{\linewidth}{lcccc}
\Xhline{1pt}
\textbf{Method} & \textbf{\# Frames} & \textbf{\begin{tabular}[c]{@{}c@{}}Trainable\\ Params (M)\end{tabular}} & \textbf{\begin{tabular}[c]{@{}c@{}}Total\\ Params (M)\end{tabular}} & \textbf{\begin{tabular}[c]{@{}c@{}}GFlops\\(Inference)\end{tabular}} \\

\Xhline{1pt}
X3D & \multirow{3}{*}{16} & 3.08 & 3.08 & 10.04 \\
MViT (B) &  & 36.34 & 36.34 & 141.8 \\
MViTv2 (S) &  & 34.27 & 34.27 & 128.64 \\\hdashline
VideoResNet & \multirow{6}{*}{32} & 33.19 & 33.19 & 25.44 \\
SlowFast &  & 34.6 & 34.6 & 1970 \\
VideoSwin (S) &  & 49.55 & 49.55 & 244.66 \\
TimeSformer &  & 121.4 & 121.4 & 1519.3 \\
ViVit &  & 88.69 & 88.69 & 906.52 \\
Ours &  & 143.06 & 306.11 & 610.4 \\\Xhline{1pt}
\end{tabularx}
\end{table}

\section{Ablation Experiments}\label{sec:ablation}
We present an ablation study to assess the contributions of the individual components within our proposed method. In addition to identifying the optimal RGB feature extractor and determining the best configuration values for the transformers used in the architecture, we compare various approaches for incorporating hierarchical information. We also analyze the influence of the number of past actions and the inclusion of location data within the contextual information. Furthermore, we investigate the impact of rephrasing with GPT-3.5 and Llama~3 on enriching contextual data, as well as the effect of different text encoders. Unless specified otherwise, all experiments in this section are conducted using the best configuration settings, which result in a $54.95\%$ top-$1$ accuracy, and we report the outcomes on fine-grained action recognition.

\subsection{Contextual Data Analysis}
The best performing model considers the last 5 actions and the location provided by fixed cameras as contextual data. Text embeddings are obtained using DistilBERT~\cite{sanh2019distilbert} for feature extraction and the \emph{prompt} used to generate the descriptions is a fixed template. Table~\ref{tab:text-features} compares the use of DistilBERT and BERT as feature extractors as well as the use of rephrasing to enhance textual descriptions exploiting two well known large language models: GPT3.5 and Llama 3.

\begin{table}[t]
\centering
\caption{Effect of rephrasing and comparison of text feature extractors.}
\label{tab:text-features}

\begin{tabular}{llcc}
\Xhline{1pt}
{\bf Feature Extractor} & {\bf Rephrasing} & {\bf Top-1} & {\bf Top-5} \\\Xhline{1pt}
\multirow{3}{*}{BERT} & GPT 3.5 & 39.46 & 73.14 \\
 & Llama 3 & \underline{40.79} & 75.18 \\
 & - & 40.53 & 75.36 \\\hdashline
\multirow{3}{*}{DistilBERT} & GPT 3.5 & 48.50 & 80.47 \\
 & Llama 3 & 49.20 & 79.84 \\
 & - & \textbf{54.95} & 84.11 \\\Xhline{1pt}
\end{tabular}
\end{table}

\begin{table}[t]
\centering
\caption{Study on the number of past actions and effect of location on contextual data.}
\label{tab:num-actions}

\begin{tabular}{llcc}
\Xhline{1pt}
{\bf \# Past Actions} & {\bf Location} & {\bf Top-1} & {\bf Top-5} \\\Xhline{1pt}
\multirow{2}{*}{1} & - & 49.17 & 82.33 \\
 & \checkmark & \underline{50.87} & 84.25 \\\hline
\multirow{2}{*}{3} & - & \underline{52.58} & 84.29 \\
 & \checkmark & 51.80 & 84.70 \\\hline
\multirow{2}{*}{5} & - & 53.46 & 84.18 \\
 & \checkmark & \textbf{54.95} & 84.11 \\\hline
\multirow{2}{*}{7} & - & \underline{52.46} & 82.33 \\
 & \checkmark & 50.95 & 80.81 \\\Xhline{1pt}
\end{tabular}
\end{table}

Results indicate that DistilBERT outperforms its predecessor for feature extraction. Additionally, a notable improvement is observed when using DistilBERT with a fixed template, as opposed to rephrasing for enhancing textual descriptions. 

Based on the optimal configuration, we conducted a study to determine the best number of past actions and assess the impact of location on contextual descriptions. The results, presented in Table~\ref{tab:num-actions}, suggest that using five past actions yields the best performance on the Hierarchical TSU dataset. Regarding location information, we observe that it provides improvements only when using 1 or 5 past actions, with variability in other cases.

\subsection{Strategies for Hierarchical Action Recognition}
As discussed in Section~\ref{sec:method}, fine-grained action recognition is performed using fused embeddings that combine both visual and contextual information. In addition, we compare different strategies for integrating fine-grained and coarse-grained action recognition into a joint learning framework. Table~\ref{tab:h-strategies} presents the results of four mechanisms, each depending on the features used to disambiguate fine- and coarse-grained information. Specifically, the strategies are: (1) using only contextual information for coarse-grained action recognition; (2) using two separate fusion transformers, employing the same visual and contextual features as the fine-grained classifier; (3) using two separate classifiers from the same fused embeddings for fine- and coarse-grained classification; and (4) sharing part of the classifier layers, while separating only the final classification layers.

\begin{table}[t]
\centering
\caption{Comparison of strategies for hierarchical action recognition.}
\label{tab:h-strategies}

\begin{tabular}{lcccc}
\Xhline{1pt}
\multirow{2}{*}{{\bf Method}} & \multicolumn{2}{c}{{\bf Fine-Grained}} & \multicolumn{2}{c}{{\bf Coarse-Grained}} \\
& {\bf Top-1} & {\bf Top-5} & {\bf Top-1} & {\bf Top-5} \\\Xhline{1pt}
(1) Contextual Data  & \textbf{54.95} & 84.11 & 73.77 & 99.15 \\
(2) Separate Fusion  & 51.83 & 82.14 & 73.10 & 99.19 \\
(3) Separate Classifier  & 51.24 & 81.88 & 74.21 & 99.11 \\
(4) Shared Classifier & 48.17 & 78.55 & 73.18 & 99.44 \\\Xhline{1pt}
\end{tabular}
\end{table}

\begin{table}[t]
\centering
\caption{Comparison of fusion strategies.}
\label{tab:data-fusion}

\begin{tabular}{lcc}
\Xhline{1pt}
{\bf Method} & {\bf Top-1} & {\bf Top-5} \\\Xhline{1pt}
(1) Separate Modalities & 49.94 & 80.03 \\
(2) Visual Concat & \textbf{54.95} & 84.11 \\
(3) Concat & 40.53 & 74.25 \\\Xhline{1pt}
\end{tabular}
\end{table}

The results indicate that the differentiation between fine-grained and coarse-grained features plays a significant role in performance. Specifically, using only contextual information for coarse-grained action recognition~(1) achieves the best performance, with a top-1 accuracy of 54.95\% for fine-grained actions and competitive results for coarse-grained actions. This suggests that contextual data is particularly effective for coarse-grained recognition and also helps to improve fine-grained action accuracy.

Although the use of contextual data improves coarse-grained recognition, it does not achieve the highest top-1 accuracy. However, it remains competitive compared to other approaches. We attribute this to the role of coarse-grained actions as an auxiliary modality that primarily enhances fine-grained performance, rather than being the main focus of the model. The highest top-1 accuracy for coarse-grained actions is achieved when incorporating visual information, as seen in the separate classifier strategy (3), which yields the best performance for coarse-grained recognition.

Furthermore, strategies (2)-(4), which involve increasing the number of shared layers between fine-grained and coarse-grained actions, lead to a reduction in fine-grained accuracy. We argue that this occurs because fine-grained actions rely on more detailed, discriminative visual features, while coarse-grained actions benefit from broader, higher-level visual cues.

\subsection{Data Fusion}
Effective fusion of data modalities is crucial for achieving strong model performance. To this end, we evaluate three different fusion strategies and conduct an ablation study on the most effective one: the fusion transformer. RGB and optical flow features are obtained separately, based on the finding that late fusion is the best strategy for the video encoder, as discussed in the next section. We assess the impact of three approaches: (1)~concatenating all three modalities (RGB, optical flow, and text), (2)~concatenating the visual modalities (RGB and optical flow) prior to the fusion transformer, and (3)~inputting the three modalities separately. Table~\ref{tab:data-fusion} contains the results, with visual concatenation prior to the fusion transformer performing best on top-$1$ accuracy.

Since the fusion transformer outperforms the concatenation approach, we conduct an ablation study on the number of attention heads and encoder layers, while keeping the embedding size fixed at 768, which corresponds to the text modality. Table~\ref{tab:fus-lay} presents the results for different numbers of encoder layers, with two layers yielding the best performance. Similarly, Table~\ref{tab:fus-heads} shows the results for varying numbers of attention heads, with two heads being the optimal configuration for our method.

\begin{table}[t]
\centering
\caption{Ablation study on the number of encoder layers on the fusion transformer.}
\label{tab:fus-lay}

\begin{tabular}{lcc}
\Xhline{1pt}
{\bf \# Enc. Layers} & {\bf Top-1} & {\bf Top-5} \\\Xhline{1pt}
1 & 54.80 & 84.81 \\
2 & \textbf{54.95} & 84.11 \\
3 & 51.32 & 81.96 \\
4 & 50.46 & 82.07 \\\Xhline{1pt}
\end{tabular}

\end{table}

\begin{table}[t]
\centering
\caption{Ablation study on the number of attention heads on the fusion transformer.}
\label{tab:fus-heads}

\begin{tabular}{lcc}
\Xhline{1pt}
{\bf \# Att. Heads} & {\bf Top-1} & {\bf Top-5} \\\Xhline{1pt}
1 & 53.21 & 83.70 \\
2 & \textbf{54.95} & 84.11 \\
4 & 52.02 & 83.22 \\
6 & 51.32 & 82.81 \\
8 & 53.61 & 83.70 \\\Xhline{1pt}
\end{tabular}
\end{table}

\subsection{Video Transformer}

We determine the optimal configuration for the video encoder by evaluating several factors: input length, RGB feature extractor, number of layers and attention heads, fusion strategy, embedding size, and the use of positional encoding and CLS tokens.

\vspace*{0.2cm}\noindent\textbf{Input Length:} In theory, longer input sequences should improve action recognition accuracy, provided there is no data loss due to excessively long sequences. Table~\ref{tab:input-blocks} shows the results of our experiments with different sequence lengths. Our findings indicate that an input length of 6.4 seconds (32 input blocks) yields the best performance. Longer sequences tend to miss specific actions within the dataset, while shorter sequences lack the temporal context needed for accurate recognition.

\vspace*{0.2cm}\noindent\textbf{RGB Feature Extractor:} Since transformer models for feature extraction are typically trained on static images rather than videos, they may encounter challenges such as motion blur caused by moving objects or people. To address this, we experiment with ViT-H/14~\cite{dosovitskiy2020image} as well as RGB feature extractors from the two-stream network TSN~\cite{TSN2016ECCV}, both using pre-trained weights from the Kinetics dataset~\cite{8099985}. Specifically, we evaluate Inception v3~\cite{szegedy2016rethinking} and BN-Inception~\cite{ioffe2015batch}. As shown in Table~\ref{tab:rgb}, despite being trained on static images, the transformer-based method achieves the best accuracy.

\vspace*{0.2cm}\noindent\textbf{Positional Information:} As pointed out in Section~\ref{sec:method}, extracted frame features lack order information. Table~\ref{tab:pos-enc} contains an ablation study on using fixed and learnable position encodings against not using them. Results show that positional encoding to preserve the order of the frames is necessary, and that learned positional encoding performs best.

\begin{table}[t]
\centering
\caption{Ablation study on the number of input blocks.}
\label{tab:input-blocks}
\begin{tabular}{cccc}
\Xhline{1pt}
{\bf \# Input Blocks} & {\bf Video duration (sec)} & {\bf Top-1} & {\bf Top-5} \\\Xhline{1pt}
8            & 1.8 & 49.45 & 80.56 \\
16           & 3.2 &  51.14 & 81.02 \\
32           & 6.4 & \textbf{54.95} & 84.11 \\
64           & 12.8 & 50.70 & 84.50 \\\Xhline{1pt}
\end{tabular}
\end{table}

\begin{table}[t]
\caption{Ablation study on RGB features extractors.}
\label{tab:rgb}
\centering
\begin{tabular}{lcc}
\Xhline{1pt}
{\bf Feature Extractor} &  {\bf Top-1} & {\bf Top-5} \\\Xhline{1pt}
Inception v3~\cite{szegedy2016rethinking} & 47.57 & 77.81 \\
BN-Inception~\cite{ioffe2015batch} &  48.54 & 79.33 \\
ViT-H/14~\cite{dosovitskiy2020image} & \textbf{54.95} & 84.11 \\\Xhline{1pt}
\end{tabular}
\end{table}

\begin{table}[t]
\caption{Ablation study on position encoding}
\label{tab:pos-enc}
\centering
\begin{tabular}{lcc}
\Xhline{1pt}
{\bf Positional Encoding} & {\bf Top-1}                         & {\bf Top-5}                         \\\Xhline{1pt}
- & 44.31  & 76.33 \\
Fixed & 52.21 & 82.96 \\
Learnable  & \textbf{54.95} & 84.11 \\\Xhline{1pt}
\end{tabular}
\end{table}

\vspace*{0.2cm}\noindent\textbf{Number of Encoder Layers and Heads:} Finding a balance between the number of encoding layers and attention heads plays a critical role in the performance of transformer models. In Table~\ref{tab:enc-lay} results show that 4 encoding layers is the best value for this task. Similarly, Table~\ref{tab:heads} shows that the model accuracy improves when using a single attention head.

\begin{table}[t]

    \centering
    \caption{Ablation study on the number of encoder layers.}
    
    \label{tab:enc-lay}
    \footnotesize
    \setlength{\tabcolsep}{3pt}
\begin{tabular}{lcc}
     \Xhline{1pt}
    {\bf \# Enc. Layers} & {\bf Top-1}                         & {\bf Top-5}                         \\\Xhline{1pt}
    2            & 52.76                         & 82.07                        \\
    4           & \textbf{54.95}                 & 84.11                          \\
    6           & 54.50 & 84.18 \\
    8           & 54.76 & 83.03 \\\Xhline{1pt}
    \end{tabular}
\end{table}

\begin{table}[t]
\centering
    \caption{Ablation study on the number of attention heads.}
    \label{tab:heads}
    \footnotesize
    \setlength{\tabcolsep}{3pt}
    \begin{tabular}{ccc}
    \Xhline{1pt}
{\bf \# Att. Heads} & {\bf Top-1}          & {\bf Top-5}                         \\\Xhline{1pt}
1            & \textbf{54.95} & 84.11 \\
2           & 52.02 & 81.92 \\
4           & 44.98 & 76.58 \\
8           & 42.91 & 74.84 \\
16           & 42.65 & 75.84 \\\Xhline{1pt}
    \end{tabular}

\end{table}

\vspace*{0.2cm}\noindent\textbf{Embedding Size and Visual Fusion Strategy:} We experiment with three different embedding dimensions on two visual fusion strategies: early and late. Early fusion involves concatenating feature vectors before the input to the transformer encoder. In contrast, a late fusion strategy requires two video encoders, one for RGB features and one for optical flow features. The results are shown in Table~\ref{tab:emd-dim}, with late fusion and an embedding size of 2048 providing the best top-$1$ accuracy.

\begin{table}[t]
\centering
\caption{Ablation study on the embedding dimension and visual fusion strategies.}
\label{tab:emd-dim}

\begin{tabular}{lccc}
\Xhline{1pt}
{\bf Fusion Type}        & {\bf Embedding Size} & {\bf Top-1} & {\bf Top-5} \\\Xhline{1pt}
\multirow{3}{*}{Early} & 768 & 43.98 & 73.18 \\
 & 1024 & 44.35 & 75.81 \\
 & 2048 & \underline{53.72} & 83.96 \\ \hline
\multirow{3}{*}{Late} & 768 & 45.35 & 77.36 \\
 & 1024 & 45.94 & 76.47 \\
 & 2048 & \textbf{54.95} & 84.11 \\\Xhline{1pt}
\end{tabular}
\end{table}

\vspace*{0.2cm}\noindent\textbf{Encoder Representations:} The output of the transformer encoder provides a representation for each input token. To avoid biasing the model towards a particular video block, we experiment with two different representations as explained in Section~\ref{sec:method}. As shown in Table~\ref{tab:cls-tkn}, using a class token improves the encoder's representation, resulting in a more robust model compared to the mean of each of the encoder's output tokens.

\begin{table}[t]
\centering
\caption{Ablation study on the encoder representation.}

\label{tab:cls-tkn}
\begin{tabular}{lcc}
\Xhline{1pt}
{\bf Encoder Representation} & {\bf Top-1}                         & {\bf Top-5}                         \\\Xhline{1pt}
CLS token            & \textbf{54.95}                 & 84.11 \\
Mean           & 45.72                & 78.21                                             \\\Xhline{1pt}
\end{tabular}
\end{table}

\clearpage
\section{Conclusion}\label{sec:conc}
In this work, we proposed a method that effectively leverages both contextual information from past actions, location and domain information, and the hierarchical structure of actions, to improve action recognition accuracy.

To fully leverage these contextual and structural cues, we propose the Hierarchical TSU dataset, an extension of the original TSU with a two-level annotation hierarchy and enriched contextual labels. This dataset supports a more comprehensive evaluation of hierarchical and context-aware action recognition methods.

Through extensive experimentation on the proposed Hierarchical TSU dataset, as well as two additional benchmarks, IkeaASM and Assembly101, we demonstrate that long-term contextual information can be successfully aggregated using language models. In addition, when available, contextual cues such as environment and current location, if available, further contribute to improving action recognition performance.

Hierarchical structures also lead to improvements, although their impact is generally smaller compared to contextual information. Our results highlight the importance of adequately structuring action annotations to fully exploit hierarchical relationships.

\section*{Acknowledgment}
\sloppy
This work has been funded by the Valencian regional government CIAICO/2022/132 Consolidated group project AI4Health, and by the Spanish State Research Agency (AEI) and ERDF/EU under grant: GEMELIA PID2024-161711OB-I00. This work is also a part of the ENIA Chair of Artificial Intelligence from the University of Alicante (TSI100927-2023-6) funded by the Recovery, Transformation and Resilience Plan from the European Union Next Generation through the Ministry for Digital Transformation and the Civil Service. This work has also been supported by a Spanish national and two regional grants for PhD studies , FPU21/00414, CIACIF/2021/430 and CIACIF/2022/175.

\bibliographystyle{IEEEtran}
\bibliography{biblio}

\end{document}